\documentclass{article}
\usepackage{ijcai25}

\usepackage{times}
\usepackage{soul}
\usepackage{url}
\usepackage[hidelinks]{hyperref}
\usepackage[utf8]{inputenc}
\usepackage[small]{caption}
\usepackage{graphicx}
\usepackage{amsmath}
\usepackage{amsthm}
\usepackage{booktabs}
\usepackage{algorithm}
\usepackage{algorithmic}
\usepackage[switch]{lineno}
\usepackage{soul, color}
\usepackage{changes}
\usepackage{hyperref}
\usepackage{amssymb}
\usepackage{multirow}
\usepackage{makecell}
\usepackage{stfloats}

\title{VidLBEval: Benchmarking and Mitigating Language Bias\\in Video-Involved LVLMs}

\author{
Yiming Yang$^1$
\and
Yangyang Guo$^2$
\and
Hui Lu$^1$\and
Yan Wang$^3$\\
\affiliations
$^1$Nanyang Technological University\\
$^2$National University of Singapore\\
$^3$Sichuan University\\
\emails
\{yiming014, hui007\}@e.ntu.edu.sg,
guoyang.eric@gmail.com,
wangyanscu@hotmail.com
}

\begin{document}

\maketitle

\begin{abstract}
Recently, Large Vision-Language Models (LVLMs) have made significant strides across diverse multimodal tasks and benchmarks.
This paper reveals a largely under-explored problem from existing video-involved LVLMs - \textbf{language bias}, where models tend to prioritize language over video and thus result in incorrect responses.
To address this research gap, we first collect a Video Language Bias Evaluation Benchmark, which is specifically designed to assess the language bias in video-involved LVLMs through two key tasks: \textit{ambiguous video contrast} and \textit{interrogative question probing}. Accordingly, we design accompanied evaluation metrics that aim to penalize LVLMs being biased by language.
In addition, we also propose Multi-branch Contrastive Decoding (MCD), introducing two expert branches to simultaneously counteract language bias potentially generated by the amateur text-only branch. Our experiments demonstrate that i) existing video-involved LVLMs, including both proprietary and open-sourced, are largely limited by the language bias problem; ii) our MCD can effectively mitigate this issue and maintain general-purpose capabilities in various video-involved LVLMs without any additional retraining or alteration to model architectures.
\end{abstract}

\section{Introduction}

Building on the significant advancements of Large Language Models (LLMs)~\cite{achiam2023gpt,dubey2024llama,yang2024qwen2}, Large Vision-Language Models (LVLMs) have recently garnered considerable attention~\cite{li2023blip,zhu2023minigpt,chen2024internvl}. Representative models such as LLaVA~\cite{liu2024improved} and Video-ChatGPT~\cite{maaz2023video} exhibit impressive capabilities across a variety of multimodal tasks and their associated benchmarks. However, despite their potential, LVLMs have suffered from a \textbf{language bias problem}, which often manifests as skewed shortcuts between questions and responses.

Some previous studies attribute the cause of this issue to mismatched model sizes between the base LLM and vision encoder within LVLMs~\cite{rohrbach2018object:chairs,chen2024internvl}. In particular, the involved language model size is often ten times larger than the vision encoder, leading to a tendency to prioritize language over vision~\cite{guan2024hallusionbench,leng2024mitigating:vcd,liu2023mitigating:lrv}. To expose this problem in image-based LVLMs, HallusionBench~\cite{guan2024hallusionbench} and AutoHallusion~\cite{wu2024autohallusion} perturb each input instance by removing or editing the given image, and then probe the potentially contradictory responses of 
\(<\)original image, perturbed image\(>\) pairs.
In addition, some methods aim to address this issue by contrasting distorted visual inputs~\cite{leng2024mitigating:vcd} or deliberately increasing the attention weights assigned to image tokens~\cite{liu2024paying:pai}.

\begin{figure*}[ht]
\centering
\includegraphics[width=0.95\textwidth]{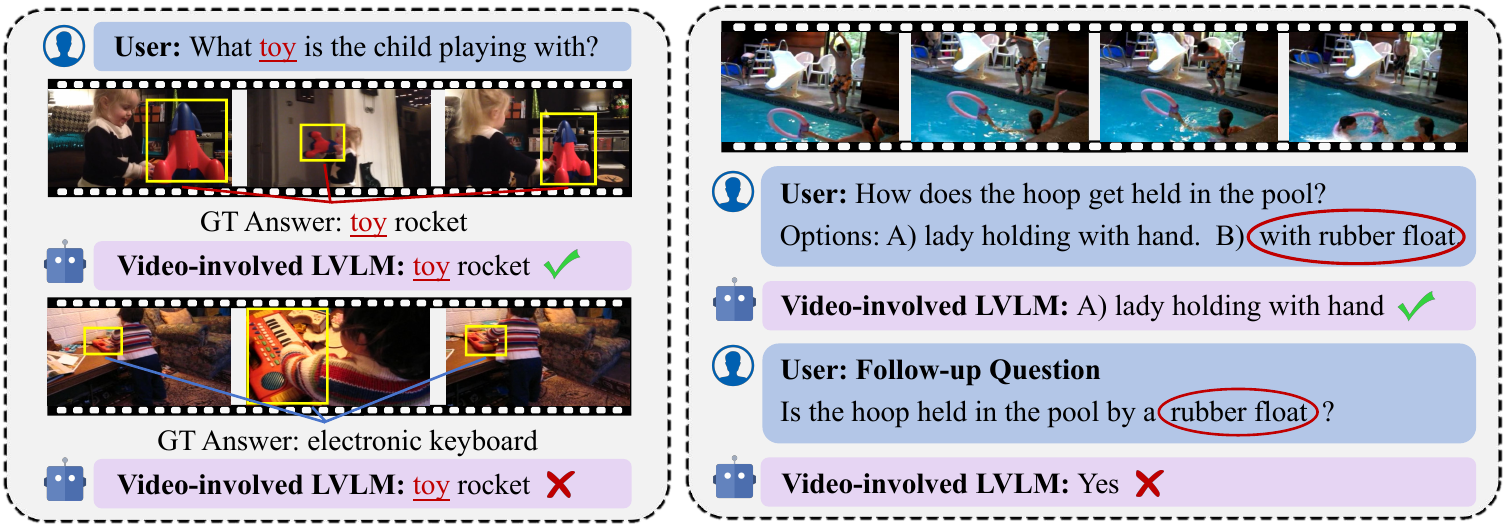}
\caption{Examples of the two involved evaluation tasks in VidLBEval. (Left) \textbf{Ambiguous Video Contrast}: We collect a complementary video that is semantically similar to the original video, yet with different answers. The LVLM provides the same answer for the same query pertaining to the two videos. (Right) \textbf{Interrogative Question Probing}: The follow-up question requires a joint understanding of the video and text. The model tends to ignore the video context by reasoning with its LLM parametric knowledge, e.g., linking hoop with rubber float.}
\label{fig:task_examples}
\end{figure*}

We note that current studies primarily focus on language bias in image-only LVLMs. This problem, however, has been largely ignored by the existing literature within the video-involved LVLMs domain. As a result, the practical video-centric applications of LVLMs, such as autonomous driving and security surveillance, are significantly compromised. To address this research gap, we first collect a \textbf{Vid}eo \textbf{L}anguage \textbf{B}ias \textbf{Eval}uation Benchmark (\textbf{VidLBEval}) to evaluate the language bias problem in video-involved LVLMs. Our VidLBEval involves two evaluation tasks: \textbf{A}mbiguous \textbf{V}ideo \textbf{C}ontrast (\textbf{AVC}) and \textbf{I}nterrogative \textbf{Q}uestion \textbf{P}robing (\textbf{IQP}). For the former, we pair each original video with either 1) another relevant video or 2) its distorted counterpart. These instances are maintained with distinct answers, which will penalize these models that consistently respond with the same answer for the same query. For the latter, we curate follow-up questions beyond the original query to challenge the model's prediction confidence. These newly generated questions are highly grounded in the joint understanding of \emph{one original answer option} and \emph{the given video} (see Figure~\ref{fig:task_examples} for the dataset examples).

Our second contribution in this work is a multi-branch contrastive decoding method. To this end, we introduce two expert branches to simultaneously counteract the language bias potentially generated by the amateur text-only branch. 
Specifically, beyond the weak expert inheriting the original model process, we design a strong expert branch to lay more attention on video features, prioritizing the reasoning over video content. We then apply this method to three state-of-the-art video-involved LVLMs\textemdash VideoLLaVA~\cite{lin2023video:videollava}, VideoLLaMA2~\cite{cheng2024videollama:videollama2}, and VideoGPT+~\cite{maaz2024videogpt+}\textemdash and evaluate its effectiveness on the VidLBEval Benchmark. Through extensive experiments, our proposed method achieves consistent improvements in language bias reduction on all these three LVLMs, demonstrating its generalization capability. 
On the other hand, we also show that the reasoning abilities on other general-purpose benchmarks, such as SEEDBench and MVBench, are preserved to a large extent.
Additionally, our approach requires no additional retraining or alteration to the base model architectures. 

In summary, to the best of our knowledge, we are the first to construct a benchmark that is specially designed to evaluate the language bias in video-involved LVLMs\footnote{The dataset will be made available to the public.}. 
We believe that the VidLBEval benchmark, along with other video-involved datasets, will provide a more comprehensive assessment of the video understanding capabilities of existing LVLMs, and thus aid further advancements in LVLM development.

\section{Related Work}

\noindent \textbf{Language Bias in VQA.} Language bias has long been recognized as a challenging problem for conventional visual question answering (VQA). Previous methods in alleviating this problem can be roughly categorized into three groups: ensemble learning, contrastive learning, and loss re-scaling. Approaches in the first group~\cite{cadene2019rubi:rubi,clark2019don} introduce an additional bias branch which is trained with the original input in an ensemble manner. Contrastive learning-based debiasing methods~\cite{liang2020learning:97,si2022towards:137} first generate positive and negative samples using data augmentation techniques. These samples are then utilized to jointly optimize the model with a contrastive learning loss alongside the original classification loss. The last group methods~\cite{ijcai2021p98:ijcai,wu2019self:scr} address this problem with inspiration from class-imbalance mechanisms. To this end, each instance-aware loss is re-weighted based on training data statistics to achieve fair training. 

\noindent \textbf{Hallucination in LVLMs.}
Hallucination in LVLMs often refers that the generated textual responses are plausible but contradictory to the associated visual content~\cite{zhou2023analyzing:137,calibrated2024}. Some initial efforts have been devoted to building benchmarks to probe the hallucinatory level of LVLMs. For instance, CHAIR~\cite{rohrbach2018object:chairs} and GAVIE~\cite{liu2023mitigating:lrv} instruct models to generate a free-form caption to reveal their exposure to errors, POPE~\cite{li2023evaluating:pope}, HallusionBench~\cite{guan2024hallusionbench} and AutoHallusion~\cite{wu2024autohallusion} query models in terms of visual reasoning aspects with binary questions. Besides, hallucination mitigation has also attracted extensive interest recently. Some data augmentation methods like LRV-Instruction~\cite{liu2023mitigating:lrv} and HalluciDoctor~\cite{yu2024hallucidoctor} introduce additional negative and counterfactual data to fine-tune LVLMs. Other approaches propose to leverage contrastive decoding~\cite{leng2024mitigating:vcd,liu2024paying:pai,code2024} or reinforcement learning from human feedback~\cite{gunjal2024detecting:HalDetect,yu2024rlhf} to address this problem.
Overall, hallucination in LVLMs often manifests with multiple dimensions, wherein language bias contributes a significant factor.
As a result, performing language debiasing greatly assists the reduction in hallucination, therefore improving the reliability of LVLMs.

\noindent \textbf{Benchmarks for Video-Involved LVLMs.} The pervasiveness of LVLMs is accompanied by continual development in video-involved benchmarks. SEEDBench~\cite{li2024seed:seedbench} and Video-Bench~\cite{ning2023video:videobench} cover a wide variety of video-centric tasks and aim to provide a comprehensive evaluation for video understanding capabilities. However, some studies find that these general benchmarks suffer from the static spatial bias from single frames~\cite{dblei:singlebias}. To approach this, MVBench~\cite{li2024mvbench} and Tempcompass~\cite{liu2024tempcompass} curate video instances covering more temporal aspects such as speed, moving direction, attribute change, and event order. Besides, Video-MME~\cite{fu2024videomme} collects long videos that last up to one hour in duration. Unlike these benchmarks, we propose to evaluate LVLMs from the dimension of language bias, which we believe, constitutes an essential component for video understanding yet received no attention in the existing literature.

\begin{figure*}[ht]
\centering
\includegraphics[width=\textwidth]{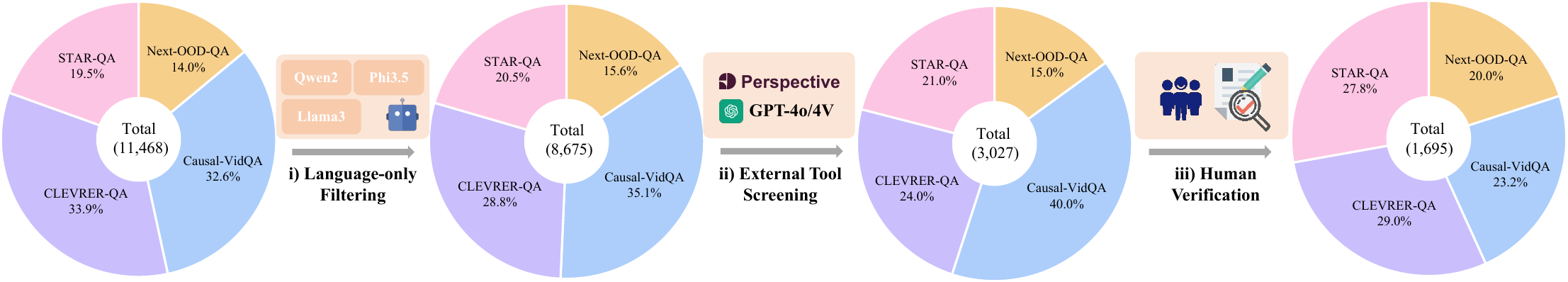}
\caption{VidLBEval quality control pipeline. i) We first filter out questions that can be answered correctly without referring to the associated video by utilizing several LLMs such as Qwen2. ii) External tools, i.e., Perspective API and GPT-4o/4V, are then employed for further safety checks. iii) Finally, we conduct human verification to review the results, leading to 1,695 high-quality samples for our VidLBEval dataset.}
\label{fig:filtering_all}
\end{figure*}

\section{VidLBEval Dataset Collection}
Our benchmark dataset is built upon four publicly available video QA datasets: Next-OOD-QA~\cite{zhang2023next:nextood}, Causal-VidQA~\cite{li2022representation:Causal-VidQA}, CLEVRER-QA~\cite{yi2019clevrer}, and STAR-QA~\cite{wu2024star:starqa}. We source these datasets hinging on two criteria: 1) The validation or test sets have rarely been used in LVLM pre-training, which prevents the potential data leakage problem. 2) The datasets are enabled to cover a broad range of video concepts such as \textit{movement direction} and scenarios like \textit{action count}.
With these anchor 
\(<\)video, question, answer options\(>\) candidates, we then construct our VidLBEval dataset.

\subsection{Ambiguous Video Contrast (AVC)}

Our first evaluation task queries LVLMs with the same question and different videos sharing similar semantics, where the answers are distinct. The motivation is that a highly biased model is prone to predict the same answer for the same question irrespective of the video context.

We implement this idea with a two-stage data construction pipeline. 
To facilitate the evaluation as well as increase the task difficulty, we first employ the VideoMAE ~\cite{tong2022videomae} model to extract visual features of all videos, based on which the pair-wise cosine similarity among videos is calculated. For each original video, we retrieve the most similar video that resembles highly in semantic content. In addition, we obtain another distorted video as a complement with the aid of applying Gaussian noise. 
In the second stage, we leverage GPT-4V~\cite{gpt4v} to generate detailed video descriptions. By the combination of the descriptions of the two paired videos, we then instruct GPT-4 to formulate a question that applies to both videos\footnote{We do not generate questions and answers directly from videos with GPT-4V as it leads to more hallucinations.}. 
In particular, the ground-truth answer to each given question is complemented with multiple distracted answers. 
These distractors are ensured to maintain a high similarity with the ground-truth one in semantics.

\subsection{Interrogative Question Probing (IQP)}

In addition to the first task using distracted videos, we also explore complementing the original question with follow-up questions. 
Our intention for this evaluation task is two-fold: 1) the newly introduced questions require joint video-text understanding and 2) the models are expected to maintain great consistency of the original answering and its follow-up process. To this end, we first concatenate the original question with both the ground-truth answer and other candidate options. The combined text is then prompted to GPT-4 for generating follow-up questions with binary answers. These questions are crafted to introduce misleading information to challenge the model's consistency.

We follow existing studies~\cite{yinyang} to focus on binary questions for two specific reasons.
First, answering binary questions is generally easier than open-ended ones, and can be seen as a second visual concept perception verification. 
Second, existing LVLMs are shown to deliver affirmative responses regardless of the visual context~\cite{li2023evaluating:pope}. We therefore, keep a balanced distribution of yes and no answers to eliminate such a visual priming bias.

\subsection{Quality Control \& Data Statistics}

We show the quality control pipeline and the output from each step in Figure~\ref{fig:filtering_all}.

\noindent \textbf{Language-only Filtering.} We expect the collected VidLBEval dataset to require grounded visual reasoning, which cannot be addressed with the question only~\cite{mmstar2024}. To this end, we input the generated questions into three powerful LLMs, including Llama3~\cite{dubey2024llama}, Qwen2~\cite{yang2024qwen2}, and Phi3.5~\cite{abdin2024phi}. The questions which can be blindly answered without looking at the associated video are thereafter filtered out.

\noindent \textbf{External Tool Screening.} We then utilize the Perspective API\footnote{\href{https://developers.perspectiveapi.com/}{https://developers.perspectiveapi.com/}.} to assess the potential negative impacts of the generated sentences, such as rudeness and toxic content. In addition, GPT-4V serves as an expert for the AVC task to avoid the cases that some questions become unanswerable~\cite{unkvqa}. GPT-4o, in the IQP task, evaluates the generated questions on various aspects such as logical coherence and lexical precision, with samples passing the threshold kept.

\noindent \textbf{Human Verification.} We finally involve further human verification, which results in our VidLBEval dataset with 521 and 1,174 multiple-choice QA pairs for the AVC task and IQP task, respectively. The comparisons between our final VidLBEval dataset and other related datasets are shown in Table~\ref{tab:comparison}.


\begin{table*}[ht]\scriptsize
    \begin{minipage}{0.62\textwidth} 
        \vspace{0.39cm}
        \centering
        \begin{tabular}{lcccc}
            \toprule
            \rule[0pt]{0pt}{3ex}\hspace*{2em}Datasets & Temporal & \makecell{Open\\Domain} & \makecell{Language\\Bias} & \makecell{Evaluation\\Metrics} \\
            \midrule
             {\textit{Image-only LVLM datasets}} & & & & \\
            \hspace*{2em}HallusionBench~\cite{guan2024hallusionbench} &  {\textbf{\texttimes}} &  {\checkmark} &  {\checkmark} & LLM Assessment      \\
            \rule[0pt]{0pt}{3ex}\hspace*{2em}AutoHallusion~\cite{wu2024autohallusion} &  {\textbf{\texttimes}} &  {\checkmark} &  {\checkmark} & LLM Assessment       \\
            \midrule
             {\textit{Video-involved LVLM datasets}} & & & & \\
            \hspace*{2em}SEEDBench~\cite{li2024seed:seedbench} &  {\checkmark} &  {\textbf{\texttimes}} &  {\textbf{\texttimes}} & ACC   \\
            \rule[0pt]{0pt}{3ex}\hspace*{2em}Video-Bench~\cite{ning2023video:videobench} &  {\checkmark} &  {\checkmark} &  {\textbf{\texttimes}} & ACC     \\
            \rule[0pt]{0pt}{3ex}\hspace*{2em}MVBench~\cite{li2024mvbench} &  {\checkmark} &  {\checkmark} &  {\textbf{\texttimes}} & ACC   \\
            \rule[0pt]{0pt}{3ex}\hspace*{2em}Tempcompass~\cite{liu2024tempcompass} &  {\checkmark} &  {\checkmark} &  {\textbf{\texttimes}} & ACC/LLM Assessment   \\
            \rule[0pt]{0pt}{3ex}\hspace*{2em}Video-MME~\cite{fu2024videomme} &  {\checkmark} &  {\checkmark} &  {\textbf{\texttimes}} & ACC    \\
             \rule[0pt]{0pt}{3ex}\hspace*{2em}\textbf{VidLBEval (Ours)} &  {\checkmark} &  {\checkmark} &  {\checkmark} & BVC/CR/RA \\
            \bottomrule
        \end{tabular}
        \captionof{table}{Comparison with related datasets from four aspects. In the evaluation metrics, ACC refers to Accuracy, and GPT-4 series models predominate in LLM Assessment.}
        \label{tab:comparison}
    \end{minipage}%
    \hspace{0.02\textwidth}
    \begin{minipage}{0.36\textwidth}
        \centering
        \includegraphics[width=0.95\linewidth]{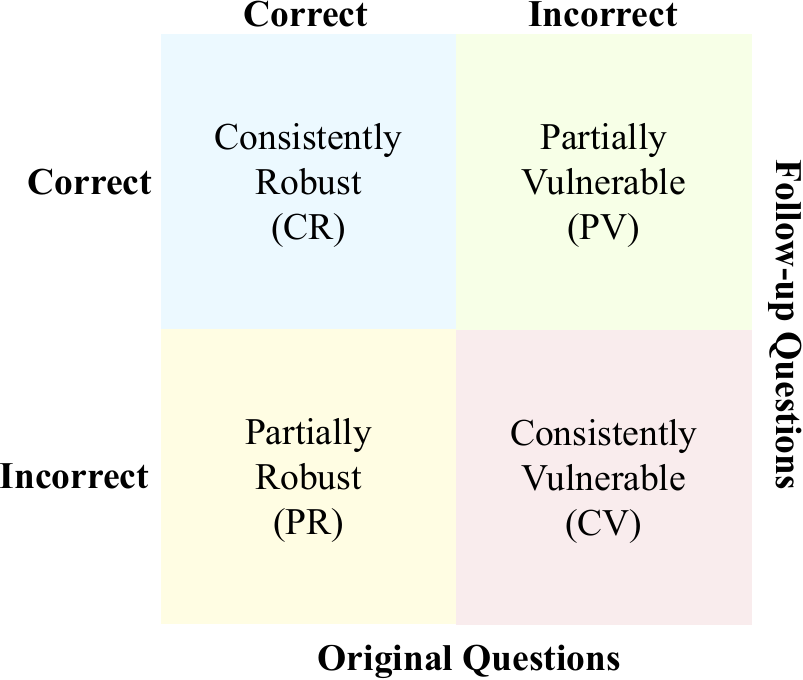}
        \captionof{figure}{Prediction interplay of the answers from the original questions and the follow-up questions.}
        \label{fig:confusion_matrix}
    \end{minipage}
\end{table*}

\subsection{Evaluation Metrics}

In the first \textbf{AVC} task, since the instances are maintained with distinct answers, for the same question, we aim to penalize LVLMs that consistently provide the same response. In particular, for each question, we consider the \textit{pairwise prediction consistency} between the original video and the related video or the distorted video. Under this context, we compute the paired instances that \textit{result in the same prediction yet at least one answer prediction is incorrect}. 
It is because we care more about the language bias effect than the answering accuracy.  
Recall the left example of Figure~\ref{fig:task_examples}, we consider the model biased when it consistently responds with \textit{toy rocket} across the two given videos.
In this way, we design a \textbf{B}iased \textbf{V}isual \textbf{C}onsistency (\textbf{BVC}) metric that accumulates these pairs over the whole dataset. We further divide the BVC into two categories: 1) $\text{BVC}_{rel}$ for the relevant video and 2) $\text{BVC}_{dis}$ for the distorted counterpart.
A lower BVC corresponds to a better model that bears less language bias.  

Moreover, for the second \textbf{IQP} task, we aim to evaluate the logical consistency of model outputs and ensure that questions are not answered through random guessing. The follow-up questions are designed to be logically correlated over a video. Based on this intuition, we introduce \textbf{T}ext \textbf{C}onsistency \textbf{R}ate (\textbf{TCR}) and \textbf{R}obust \textbf{A}ccuracy (\textbf{RA}) metrics. We first classify the answering prediction attributes in Figure~\ref{fig:confusion_matrix} and then use $N$ as a general notation for the sample count in the respective category. Based on this figure, the \textbf{TCR} is defined as \( TCR = N_{CR} / (N_{CR} + N_{PR}) \), while the \textbf{RA} is given by \( RA = N_{CR} / (N_{CR} + N_{PR} + N_{PV} + N_{CV}) \).

Beyond conventional accuracy metrics, our novel evaluation strategy provides a more effective mechanism for language bias probing while eliminating dependency on external LLMs. Specifically, LLM assessment introduces additional costs and variability, as open-source models may produce inconsistent outputs across different versions for the same input, such as GPT-4 and GPT-4V-Turbo used by~\cite{guan2024hallusionbench} and~\cite{wu2024autohallusion}, respectively. In contrast, our metrics are designed to be both stable and deterministic,
ensuring a consistent and cost-efficient evaluation protocol.

\begin{table*}[ht]
    \centering
    \resizebox{0.91\textwidth}{!}{%
    \begin{tabular}{lrrcccccccc}
        \toprule
        \hspace*{2em}\multirow{2}{*}{Model} & \multirow{2}{*}{\#Param} & \multirow{2}{*}{\#Frame} & & \multicolumn{4}{c}{AVC} & & \multicolumn{2}{c}{IQP} \\
        \cmidrule{5-8} \cmidrule{10-11}
        & & & & $\text{ACC}_{rel} \uparrow$ & $\text{BVC}_{rel} \downarrow$ & $\text{ACC}_{dis} \uparrow$ & $\text{BVC}_{dis} \downarrow$ & & TCR $\uparrow$ & RA $\uparrow$ \\
        \midrule
         {\textit{Open-Source}} & & & & & & & & & \\
        \hspace*{2em}VideoChat~\cite{videochat} & 7B & 8 & & 11.90 & 34.64 & 17.27 & 21.11 & & 28.36 & 8.09 \\
        \hspace*{2em}Video-ChatGPT~\cite{maaz2023video} & 7B & 100 & & 4.03 & 55.00 & 11.52 & 48.37 & & 28.18 & 10.39 \\
        \hspace*{2em}VideoLLaVA~\cite{lin2023video:videollava} & 7B & 8 & & 53.55 & 47.11 & 63.15 & 69.79 & & 17.66 & 7.58 \\
        \hspace*{2em}VideoChat2~\cite{li2024mvbench} & 7B & 16 & & 15.74 & \textbf{29.38} & 26.87 & 78.74 & & 11.33 & 6.47 \\
        \hspace*{2em}LLaVA-NeXT~\cite{llava-next} & 7B & 32 & & 51.35 & 40.32 & 64.23 & 69.35 & & 14.16 & 7.12 \\ \hspace*{2em}VideoLLama2~\cite{cheng2024videollama:videollama2} & 7B & 8 & & 72.55 & 46.15 & 81.57 & 61.46 & & 28.89 & 10.31 \\
        \hspace*{2em}VideoGPT+~\cite{maaz2024videogpt+} & 3.8B & 16 & & 73.32 & 33.09 & 80.81 & 74.00 & & \textbf{32.14} & \textbf{21.47} \\
        \midrule
         {\textit{Proprietary}} & & & & & & & & & \\
        \hspace*{2em}GPT-4V~\cite{gpt4v} & - & 10 & & \textbf{92.84} & 34.29 & \textbf{83.44} & \textbf{9.88} & & 28.55 & 17.84 \\
        \bottomrule
    \end{tabular}
    }
    \caption{Benchmark results on VidLBEval using greedy decoding. \#Param indicates the number of LLM parameters, while $\text{ACC}_{rel}$ and $\text{ACC}_{dis}$ denote the joint accuracy of \(<\)original, relevant\(>\) and \(<\)original, distorted\(>\) video pairs, respectively. The best performance is highlighted in bold.}
    \label{tab:benchmarking}
\end{table*}

\section{Method}

\subsection{Preliminaries and Motivation}
We consider a video-involved LVLM, parameterized by \( \theta \), typically designed to generate the response $y$ given a video $v$ and a textual query $x$. The operation starts by passing the video $v$ through a vision encoder, followed by a projector that maps it into a set of visual tokens. These visual tokens are then concatenated with the text tokens to serve as the input for the language model in LVLM. The response $y$ is auto-regressively generated from the probability distribution conditioned on the query $x$, the video $v$, and the generated tokens $y_{<t}$ up to the time step $t - 1$,
\begin{equation}
\begin{aligned}y_{t}&\sim p_{\theta}(y_{t}| v,x,y_{<t})\propto\operatorname{logit}_{\theta}(y_{t}|  v,x,y_{<t}),
\end{aligned}
\end{equation}
where $y_{t}$ represents the token sampled at the $t$-th time step, and $\operatorname{logit}_\theta$ refers to the predicted logits after the softmax function by model $\theta$. 

Building on the foundational advancements in LLMs~\cite{li2022contrastive:cd}, recent studies~\cite{leng2024mitigating:vcd,liu2024paying:pai,code2024} have introduced the Visual Contrastive Decoding (VCD) mechanism to enhance the visual understanding capability, thereby reducing the hallucination problem in image-based LVLMs. The next-token probability $p_{vcd}$ in VCD is generally expressed as:
\begin{align}
p_{vcd}=(1+\gamma)p_{\theta}(y_{t}|v,x,y_{<t}) - \gamma p_{\theta}(y_{t}|x,y_{<t}),
\end{align}
where $p_{\theta}(y_{t}|x,y_{<t})$ represents the amateur branch with pure textual inputs, and $\gamma$ controls the penalty extent. It is worth noting that we do not consider other alternatives for the amateur branch, such as the same model architecture with different parameters or other inputs.

\noindent \textbf{Motivation.} Our motivation for this method is two-fold. First, to the best of our knowledge, the VCD algorithm has been rarely studied in video-involved LVLMs. As such, we intend to explore its effectiveness within this specific domain. Second, we believe that video understanding requires a more nuanced approach compared to image understanding. It is because that videos inherently exhibit richer temporal dynamics than images, making it crucial to focus more on video frames. 

\subsection{Multi-branch Contrastive Decoding}

\begin{figure}[t]
\centering
\includegraphics[width=0.44\textwidth]{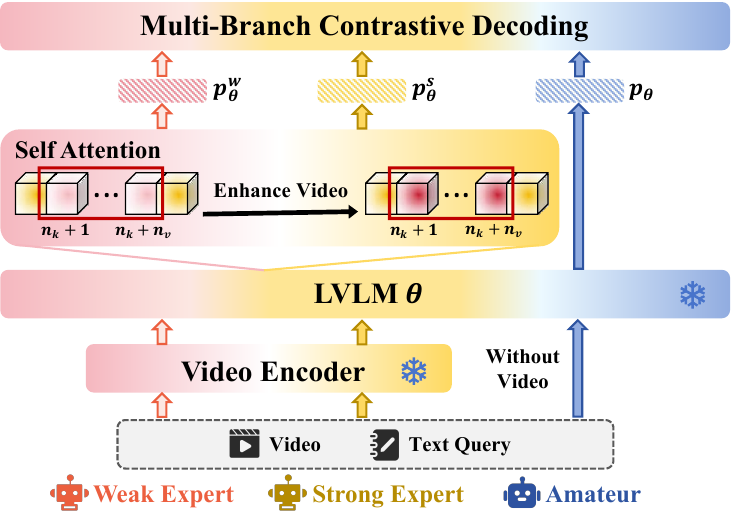}
\caption{Architecture of our MCD. Two expert branches are introduced to simultaneously mitigate language bias from the amateur text-only branch: the weak expert retaining the original model process and the strong expert laying more attention on video features.}
\label{fig:mcd}
\end{figure}

We propose a \textbf{M}ulti-branch \textbf{C}ontrastive \textbf{D}ecoding (\textbf{MCD}) framework, as shown in Figure~\ref{fig:mcd},
\begin{small}
\begin{equation}
\begin{aligned}
    p_{mcd} & = (1+\gamma)p'_{\theta}(y_{t}|v,x,y_{<t})
    - \gamma p_{\theta}(y_{t}|x,y_{<t}), \\
    \text{where}\:p'_{\theta} & = \lambda p_{\theta}^w(y_{t}|v,x,y_{<t})
+ (1-\lambda) p_{\theta}^s(y_{t}|v,x,y_{<t}).
\end{aligned}
\end{equation}
\end{small}Here, $p_{\theta}^w(y_{t}|v,x,y_{<t})$ and $p_{\theta}^s(y_{t}|v,x,y_{<t})$ represent the weak expert and video-enhanced strong expert branches, respectively. The integrated expert, denoted by $p'_{\theta}$, incorporates a weighting factor $\lambda \in [0, 1]$ to balance the contributions of the two experts.
In addition to the original weak expert with multimodal input used in previous VCD methods, we introduce the video-enhanced branch as the strong expert. The new branch places greater emphasis on video content, thereby allowing visual features to be interacted more with response generation.

The MCD objective rewards text patterns preferred by the multimodal expert branch while penalizing those favored by the amateur branch. However, this can lead to the over-penalization of text-based outputs that still align with linguistic norms and common sense. To address this, we follow~\cite{li2022contrastive:cd} to introduce an adaptive plausibility constraint based on the confidence level of the output distribution:
\begin{small}
\begin{equation}\begin{aligned}
\mathcal{V}_{\mathrm{head}} (y_{<t}) & = \{y_{t}\in\mathcal{V}: \\
p_{\theta}(y_{t}| v,x,y_{<t}) & \geq\beta\operatorname*{max}_{w}p_{\theta}(w|v,x,y_{<t})\}, \\
p_{mcd}(y_t|v,x,y_{<t}) & = 0, \, \mathrm{if} \, y_t\not\in\mathcal{V}_{\mathrm{head}} \left(y_{<t}\right),
\end{aligned}\end{equation}
\end{small}where \( \mathcal{V} \) refers to the token vocabulary and \( \beta \) controls the truncation of the next token distribution, with only tokens in \( \mathcal{V}_{\mathrm{head}} \) being considered for potential candidates. This method refines the candidate pool, effectively preventing the generation of implausible tokens and preserving the quality of the generated content.

\begin{figure*}[ht]
\centering
\includegraphics[width=\textwidth]{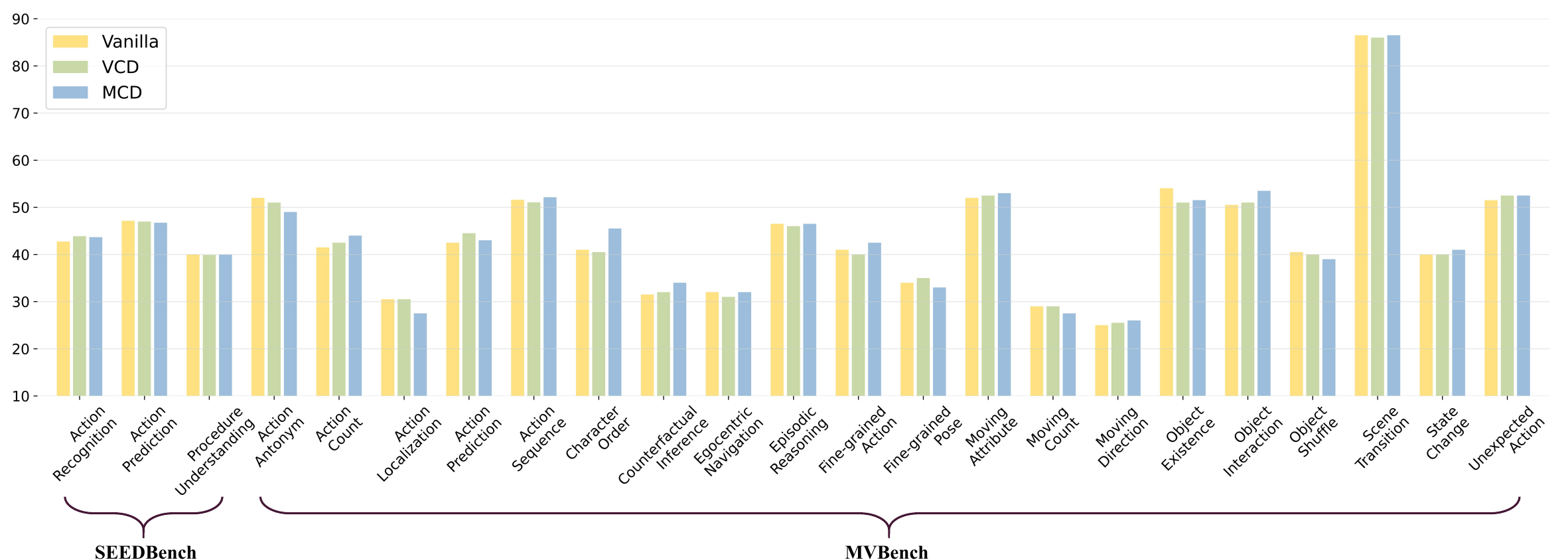}
\caption{Results on SEEDBench and MVBench when applying our proposed method to VideoLLaVA.}
\label{fig:general_dataset}
\end{figure*}

\subsection{Video-Enhanced Branch Design}

To construct the strong expert branch, we propose to increase the attention weights by updating the self-attention matrices. Specifically, we first locate the attention positions of the video tokens for the currently generated token from the attention weights $A \in \mathbb{R}^{n\times n}$ before the softmax operation, where $n$ is the sequence length. An amplification coefficient \( \alpha>=0\) is then applied to the video tokens to control the step size for generation intervention. We formulate this operation as: 
\begin{equation}
A_i=A_i+\alpha\left|A_i\right|,\,\text{where }i\in\{n_k+1,\ldots,n_k+n_v\},
\end{equation}
where $n_k$ and $n_v$ indicate the number of query tokens preceding the video token and video tokens, respectively.
Subsequently, a softmax function redistributes the attention values across all tokens. This encourages the attention mechanism to concentrate more on video information, thereby making the constructed strong expert branch video-enhanced. 

It is worth noting that the attention weights are automatically redistributed during inference, without any additional retraining or alteration to model architectures. Moreover, we maintain the same parameters for all the three branches and do not introduce any additional parameters to the LVLMs, enabling our method to be seamlessly integrated into different models without many bells and whistles.

\begin{table}[t]
    \centering
    \resizebox{0.48\textwidth}{!}{%
    \begin{tabular}{ccccccc}
        \toprule
        \multirow{2}{*}{Model} & \multirow{2}{*}{Decoding} & \multicolumn{2}{c}{AVC} & & \multicolumn{2}{c}{IQP} \\
        \cmidrule{3-4} \cmidrule{6-7}
        & & $\text{BVC}_{rel}$$\downarrow$ & $\text{BVC}_{dis}$$\downarrow$ & & TCR$\uparrow$ & RA$\uparrow$ \\
        \midrule
        \multirow{6}{*}{VLLVA}
        & Greedy & 47.11 & 69.79 & & 17.66 & 7.58 \\
        & Beam & 46.15 & 67.94 & & 11.07 & 5.03 \\
        & Nucleus & 46.22 & 71.72 & & 17.72 & 7.67 \\
        & Top-k & 46.64 & 67.35 & & 17.59 & 7.58 \\
        & VCD & 44.39 & 65.67 & & 21.50 & 9.03 \\ 
        & MCD (Ours) & \textbf{43.80} & \textbf{64.62} & & \textbf{23.18} & \textbf{9.20}\\
        \midrule
        \multirow{6}{*}{VLL2}
        & Greedy & 46.15 & 61.46 & & 28.89 & 10.31 \\
        & Beam & 45.89 & 57.00 & & 27.37 & 11.07 \\
        & Nucleus & 46.15 & 61.46 & & 27.99 & 10.56 \\
        & Top-k & 43.97 & 58.16 & & 28.64 & 10.90 \\
        & VCD & 45.21 & 62.50 & & 29.71 & 11.41 \\
        & MCD (Ours) & \textbf{40.44} & \textbf{56.67} & & \textbf{31.73} & \textbf{12.35} \\
        \midrule
        \multirow{6}{*}{VGPT+}
        & Greedy & 33.09 & 74.00 & & 32.14 & 21.47 \\
        & Beam & 30.41 & \textbf{66.20} & & 35.10 & 18.06 \\
        & Nucleus & 33.09 & 74.00 & & 32.29 & 21.55 \\
        & Top-k & 31.21 & 67.96 & & 32.30 & 21.38 \\
        & VCD & 30.66 & 70.83 & & 31.30 & 20.95 \\
        & MCD (Ours) & \textbf{29.14} & 68.18 & & \textbf{36.07} & \textbf{23.59} \\
        \bottomrule
    \end{tabular}
    }
    \caption{Results on VidLBEval when applying our proposed method to VideoLLaVA (VLLVA), VideoLLama2 (VLL2), and VideoGPT+ (VGPT+). The best performance is highlighted in bold.}
    \label{tab:mcd}
\end{table}

\section{Experiments}

\subsection{Experimental Settings}

\noindent \textbf{Baselines.}
We first benchmarked various video-involved LVLMs on VidLBEval: VideoChat (7B)~\cite{videochat}, Video-ChatGPT (7B)~\cite{maaz2023video}, VideoLLaVA (7B)~\cite{lin2023video:videollava}, VideoChat2 (7B)~\cite{li2024mvbench}, LLaVA-NeXT (7B)~\cite{llava-next}, VideoLLama2 (7B)~\cite{cheng2024videollama:videollama2}, VideoGPT+ (3.8B)~\cite{maaz2024videogpt+}, and GPT-4V~\cite{gpt4v}. Since the video interface of GPT-4V has not been released yet, we sampled 10 frames and evaluated the model using multiple images as input. On top of that, we used the default frame counts provided for other open-source models.

\noindent \textbf{Backbones.}
We applied MCD to VideoLLaVA, VideoLLama2, and VideoGPT+, which employ Vicuna 7B, Mistral 7B, and Phi-3-Mini 3.8B as language decoder, respectively. 

\noindent \textbf{Datasets.}
We utilized three datasets for detailed evaluation. Beyond our VidLBEval, SEEDBench~\cite{li2024seed:seedbench} and MVBench~\cite{li2024mvbench} serve as general-purpose benchmarks tailored to evaluate video-involved LVLMs across multiple dimensions. While they adopt accuracy as the primary evaluation metric, we introduce novel BVC, TCR, and RA metrics to provide a more effective mechanism for language bias probing. 

\noindent \textbf{Implementation Details.} 
For our proposed method, we set \(\beta\) = 0.1 and \(\gamma\) = 0.1 for all the experiments. As models vary in the lengths of their video token sequences, resulting in different levels of video neglect, we adjusted the value of \(\alpha\) and \(\lambda\) for each model to better align with its specific video sequence length. We compared our method with four regular decoding strategies: greedy decoding, beam search, nucleus sampling, and top-k sampling. 
We also included VCD strategy for comparison with our method.

\subsection{Experimental Results}
\noindent \textbf{Benchmark Results.} We summarize the overall benchmark results across eight video-involved LVLMs on VidLBEval in Table~\ref{tab:benchmarking}. Our observations are three-fold: 1) All the evaluated models consistently demonstrate severe language bias. For example, these models display a clear weak logical consistency, as evidenced by the TCR and RA remaining below one-third and one-fourth, respectively. On the other hand, the majority of $\text{BVC}_{rel}$ exceed 30\%, implying that current video-involved LVLMs confuse with similar videos. 
2) Proprietary GPT-4V shows superior results than open-source models. This discrepancy is evidently demonstrated by $\text{BVC}_{dis}$, with open-source models highlighting a marked gap at nearly 70\%.
3) IQP poses greater challenges compared to AVC. Specifically, even the best-performing models, such as GPT-4V and VideoGPT+, yield suboptimal results.

\noindent \textbf{MCD Performance on VidLBEval.} We present language bias results across three state-of-the-art video-involved LVLMs, as shown in Table~\ref{tab:mcd}. In summary, there is a notable improvement after incorporating MCD. Specifically, across various decoding settings, our MCD method consistently exceeds the baseline results by large margins.
This highlights its critical role in enhancing video-focused understanding, thereby reducing instances of language bias.

\noindent \textbf{MCD Performance on SEEDBench and MVBench.} 
In addition, we also include the evaluation of MCD on SEEDBench and MVBench to assess its impact on the general capabilities of video-involved LVLMs. With all models exhibiting comparable performance trends, we present the results of VideoLLaVA as a representative\footnote{Comprehensive results for the other two LVLMs on SEEDBench and MVBench are provided in Supplementary Material.}. As illustrated in Figure~\ref{fig:general_dataset}, MCD preserves the original general-purpose capabilities across various video-involved LVLMs.

\begin{table}[t]\small
    \centering
    \tabcolsep=5pt
    \resizebox{0.48\textwidth}{!}{%
    \begin{tabular}{ccccccccc}
        \toprule
        \multirow{2}{*}{Model} & \multirow{2}{*}{VE} & \multirow{2}{*}{OR} & & \multicolumn{2}{c}{AVC} & & \multicolumn{2}{c}{IQP} \\
        \cmidrule{5-6} \cmidrule{8-9}
        & & & & $\text{BVC}_{rel}$$\downarrow$ & $\text{BVC}_{dis}$$\downarrow$ & & TCR$\uparrow$ & RA$\uparrow$ \\
        \midrule
        \multirow{4}{*}{VLLVA} & \textbf{\texttimes} & \textbf{\texttimes} & & 47.11 & 69.79 & & 17.66 & 7.58 \\
        & \textbf{\texttimes} & \checkmark & & 44.39 & 65.67 & & 21.50 & 9.03 \\
        & \checkmark & \textbf{\texttimes} & & 44.31 & 65.45 & & 20.37 & 8.43 \\
        & \checkmark & \checkmark & & \textbf{43.80} & \textbf{64.62} & & \textbf{23.18} & \textbf{9.20} \\
        \midrule
        \multirow{4}{*}{VLL2} & \textbf{\texttimes} & \textbf{\texttimes} & & 46.15 & 61.46 & & 28.89 & 10.31 \\
        & \textbf{\texttimes} & \checkmark & & 45.21 & 62.50 & & 29.71 & 11.41 \\
        & \checkmark & \textbf{\texttimes} & & 44.76 & 60.42 & & 29.89 & 11.58 \\
        & \checkmark & \checkmark & & \textbf{40.44} & \textbf{56.67} & & \textbf{31.73} & \textbf{12.35}\\
        \midrule
        \multirow{4}{*}{VGPT+} & \textbf{\texttimes} & \textbf{\texttimes} & & 33.09 & 74.00 & & 32.14 & 21.47 \\
        & \textbf{\texttimes} & \checkmark & & 30.66 & 70.83 & & 31.30 & 20.95 \\
        & \checkmark & \textbf{\texttimes} & & 32.84 & 73.20 & & 33.59 & 22.32 \\
        & \checkmark & \checkmark & & \textbf{29.14} & \textbf{68.18} & & \textbf{36.07} & \textbf{23.59} \\
        \bottomrule
    \end{tabular}
    }
    \caption{Ablation study on video-enhanced (VE) and original branches (OR) for VideoLLaVA (VLLVA), VideoLLama2 (VLL2), and VideoGPT+ (VGPT+).}
    \label{tab:ab_branches}
\end{table}

\begin{figure}[h]
\centering
\includegraphics[width=0.44\textwidth]{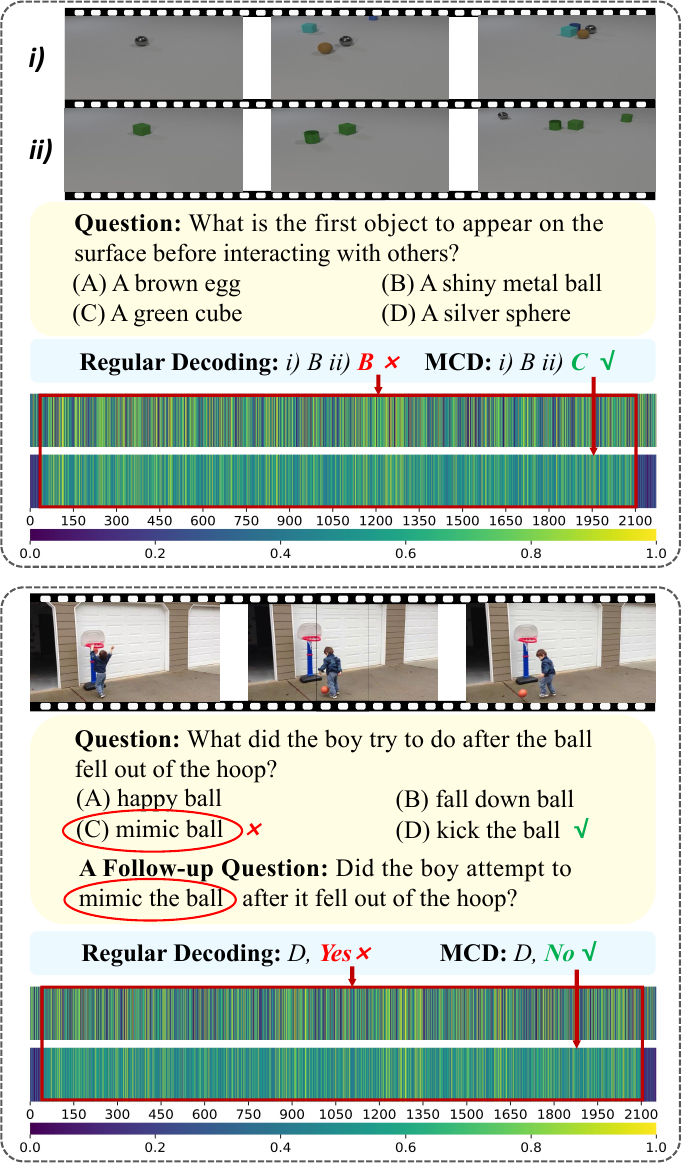}
\caption{Illustration of language bias mitigation by our proposed method on two VideoLLaVA samples from AVC and IQP. Below each sample is the corresponding visualization and comparison of the last-layer attention weights after applying our MCD method. For VideoLLaVA, the starting and ending indices of video tokens are 35 and 2098, respectively.}
\label{fig:case_study}
\end{figure}

\subsection{In-depth Analysis on MCD}
\noindent \textbf{Effect of Different Branches.}
In our proposed MCD, we introduce the video-enhanced branch, which is integrated with the original branch to generate a more robust prediction. We then evaluate the effectiveness of each respective branch, compare it against the vanilla greedy decoding, and present the results in Table~\ref{tab:ab_branches}. One can see that each branch positively contributes to the reduction of language bias. Integrating the two expert branches together delivers the best results across different models.

\noindent \textbf{Case Study on VideoLLaVA.} Figure~\ref{fig:case_study} demonstrates two cases on how regular decoding can yield language bias. In the first case, the model consistently responds with \textit{shiny metal ball}, despite \textit{green cube} being the first object presented in the other video. In the second case, the model gives the affirmative answer \textit{yes}, disregarding the fact that the boy \textit{kicked the ball} after it fell out of the hoop.
In contrast, our MCD emphasizes the video information by significantly increasing the attention weights of video tokens, thereby effectively mitigating language bias.

\section{Conclusion and Discussion}
In this paper, we address the research gap concerning language bias in video-involved LVLMs. We first introduce the VidLBEval benchmark to evaluate language bias in video-involved LVLMs, making it distinguished from existing benchmarks. Our initial findings reveal that current models suffer from severe language bias. In light of this, we propose a novel MCD approach that incorporates two expert branches to counteract the language bias potentially introduced by the text-only branch, without requiring any additional retraining or architectural modifications. Extensive experiments validate the effectiveness of MCD in reducing language bias and demonstrate its potential to enhance the overall capabilities of video-involved LVLMs.

The findings from this paper shed light on two potential future directions. 
First, it is suggested to collect extensive yet less biased data, thus aiding pretraining a fairer LVLM. 
Second, designing mitigation methods for supervised fine-tuning can act as a remedy for the inherent bias of LVLMs. 
However, striking a trade-off between general-purpose capability perservation and language bias removal is rather challenging and deserves further exploration. 


\bibliographystyle{named}
\bibliography{ijcai25}

\end{document}